\documentclass[sn-mathphys]{sn-jnl}

\usepackage{diagbox}

\jyear{2022}%

\theoremstyle{thmstyleone}%
\theoremstyle{thmstyletwo}%

\theoremstyle{thmstylethree}%

\begin{document}

\title[Article Title]{VI-PINNs: Variance-involved Physics-informed Neural Networks for Fast and Accurate Prediction of Partial Differential Equations}


\author[1]{\fnm{Bin} \sur{Shan}}\email{bin.shan@nuaa.edu.cn}

\author*[1]{\fnm{Ye} \sur{Li}}\email{yeli20@nuaa.edu.cn}

\author[1]{\fnm{Sheng-Jun} \sur{Huang}}\email{huangsj@nuaa.edu.cn}

\affil[1]{\orgname{Nanjing University of Aeronautics and Astronautics}, \orgaddress{\city{Nanjing}, \postcode{211106}, \country{China}}}


\abstract{Although physics-informed neural networks(PINNs) have progressed a lot in many real applications recently, there remains problems to be further studied, such as achieving more accurate results, taking less training time, and quantifying the uncertainty of the predicted results. Recent advances in PINNs have indeed significantly improved the performance of PINNs in many aspects, but few have considered the effect of variance in the training process. In this work, we take into consideration the effect of variance and propose our VI-PINNs to give better predictions. We output two values in the final layer of the network to represent the predicted mean and variance respectively, and the latter is used to represent the uncertainty of the output. A modified
negative log-likelihood loss and an auxiliary task are introduced for fast and accurate training. We perform several experiments on a wide range of different problems to highlight the advantages of our approach. The results convey that our method not only gives more accurate predictions but also converges faster.}

\keywords{Deep learning, Physics-informed neural networks, Differential equations, Predictive uncertainty}



\maketitle

\section{Introduction}
\label{sec:introduction}
Physics-informed neural networks(PINNs) \cite{raissi2019physics} have been proved to be efficient and powerful to handle problems from different areas with the development of deep neural networks(DNNs) \cite{cai2022physics,sahli2020physics, mao2020physics}. By combining underlying physical information usually described by partial differential equations(PDEs) with neural networks, PINNs performed better than traditional data-driven neural networks while using fewer data.

Despite the great progress in PINNs, how to effectively improve the performance of PINNs is still an issue worth studying. Bischof and Kraus \cite{bischof2021multi} summarized that current research may be concluded into four approaches: modifying the structure, divide-and-conquer/domain decomposition, parameter initialization, and loss balancing. The results of \cite{peng2020accelerating,kim2022fast} vividly conveyed to us that modifying structure does make PINNs get more knowledge to speed up convergence. The key idea of domain decomposition is to solve subproblems defined on smaller subdomains rather than the whole domain \cite{chan1994domain,smith1997domain}, and \cite{jagtap2021extended,moseley2021finite} put into together the PINNs and domain decomposition to improve the performance of vanilla PINNs. Novel algorithms for adaptive loss balancing were proposed in \cite{bischof2021multi,wang2021understanding,mcclenny2020self}.

How to quantify the uncertainty in PINNs is another issue that needs to be studied. Bayesian NNs are one of the most appropriate methods for estimating uncertainty, where the posterior is computed based on a prior distribution and training data. The work of Yang et~al. \cite{yang2021b} is the combination of PINNs and Bayesian NNs, but compared with non-Bayesian NNs, the cost of computation is expensive. Dropout \cite{srivastava2014dropout} is another method used for uncertainty estimation. During the training of NNs, we drop some neurons at random, and it seems that we are training different neural networks, so the final result is the average of the effects of diverse networks, which can be thought of as an ensemble of different models to some extent.

In this work, we propose our simple framework named variance-involved physics-informed neural networks(VI-PINNs). We hold the hopes and ideas that (i) the estimation of uncertainty in PINNs is bound to be simple but effective. Inspired by \cite{nix1994estimating,lakshminarayanan2017simple,pestourie2020active}, we output two values in the final layer, which is the main idea of heteroscedastic neural network(HNN), to represent the mean and variance, and this is an improvement in the structure of vanilla PINNs; (ii) the uncertainty should be incorporated into the training of PINNs. We modify the negative log-likelihood loss(NLL) used in HNN to make it more suitable for PINNs, and instead of using the modified loss alone, we introduce it to the PINN loss as an auxiliary task to assist the training of vanilla loss, and this is an improvement in the loss function of vanilla PINNs. The main contributions of this paper are summarized below:
\begin{itemize}
    \item[1.] We modify the structure of vanilla PINNs to model the uncertainty in an explicit way.
    \item[2.] We modify the NLL loss function commonly used in HNN to make it suitable for the training of vanilla PINNs.
    \item[3.] We incorporate the modified loss function as an auxiliary task to help the convergence of vanilla PINNs.
\end{itemize}

The rest of the paper is organized as follows: A brief introduction to the background of PINNs, together with some related work, is presented in Section \ref{sec:background}. In Section \ref{sec:methodology}, we propose our variance-involved physics-informed neural networks(VI-PINNs). Computational results for different PDE problems are provided in Section \ref{sec:results}. Finally, Section \ref{sec:conclusion} concludes the paper.

\section{Background}
\label{sec:background}
This section briefly reviews physics-informed neural networks(PINNs) in tackling partial differential equations(PDEs). Then some recent developments in improving the training efficiency and accuracy of PINNs along with quantification of uncertainty are covered.
\subsection{Physics-Informed Neural Networks(PINNs)}
\begin{figure}[H]
\centering
\includegraphics[scale=0.15]{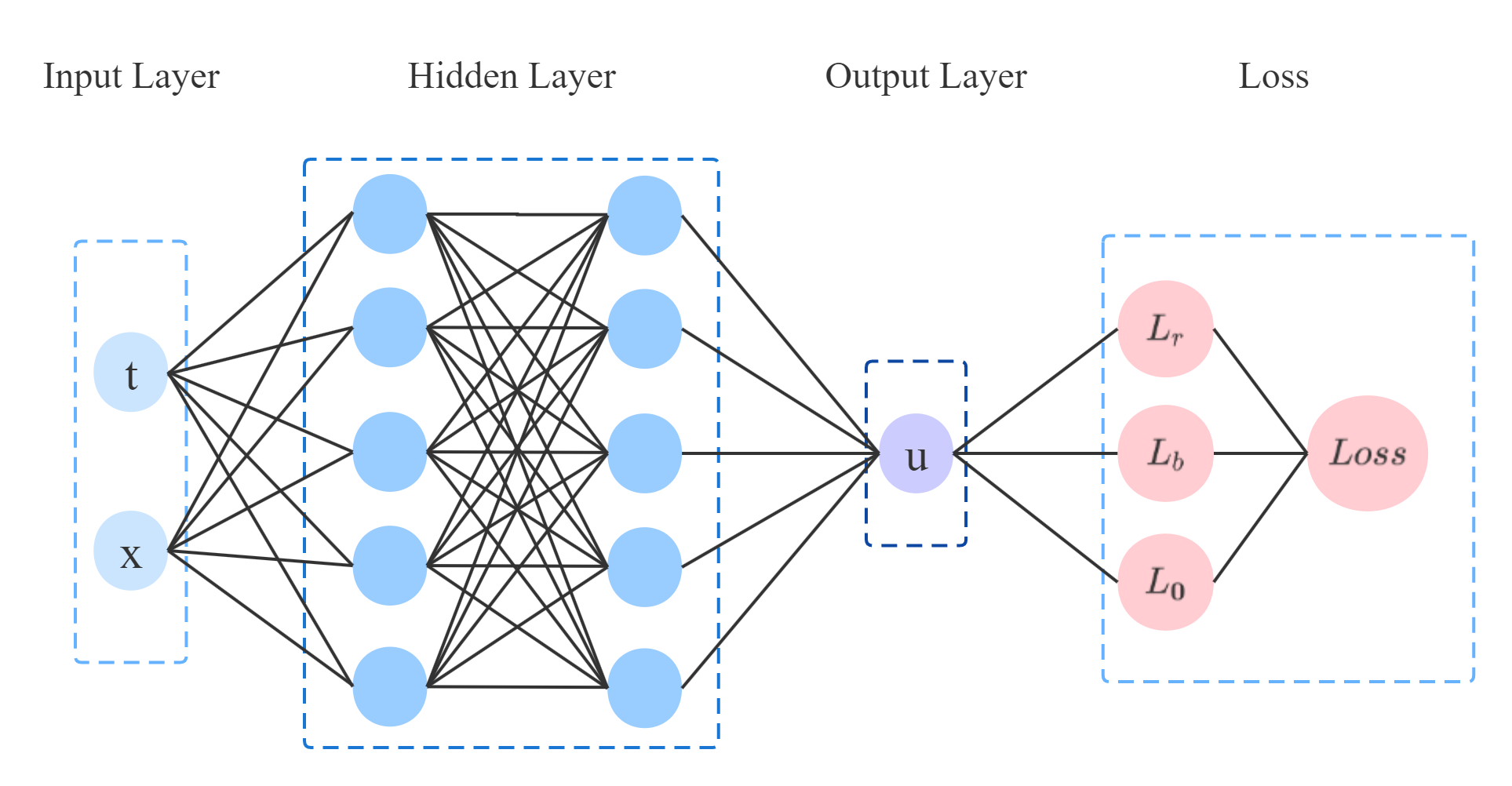}
\caption{Schematic of PINNs}
\label{fig:fig1}
\end{figure}

It has been a long time since artificial neural networks were applied to solve ordinary and partial differential equations \cite{dissanayake1994neural,lagaris1998artificial,aarts2001neural}. Recently, with the rapid development of deep learning, \cite{raissi2019physics} incorporated the differential form of the partial differential equation into the design of the loss function. Consider a partial differential equation with initial and boundary conditions taking the following general form:
\begin{equation}
\label{equ:equ1}
\mathcal{N}[u(t,x)]=f(t,x),\ x\in\Omega,\ t\in[0,T],
\end{equation}
\begin{equation}
\label{equ:equ2}
\mathcal{B}[u(t,x)]=g(t,x),\ x\in\partial{\Omega},\ t\in[0,T],
\end{equation}
\begin{equation}
\label{equ:equ3}
u(0,x)=h(x),\ x\in\Omega,
\end{equation}
where $u(t,x)$ represents the solution of PDE, $\mathcal{N}[\cdot]$ and $\mathcal{B}[\cdot]$ are linear/nonlinear differential operators, $\Omega$ is a subset of $\mathbb{R}^{D}$, and $\partial{\Omega}$ is the boundary of $\Omega$. PINNs consider approximating $u(t,x)$ by a deep neural network $u_{\theta}(t,x)$, so that the value of $\mathcal{N}[u_{\theta}(t,x)]$ and $\mathcal{B}[u_{\theta}(t,x)]$ can be computed readily and precisely using automatic differentiation \cite{baydin2018automatic,paszke2017automatic}. The residual form of Equation \eqref{equ:equ1} is defined as follows:
\begin{equation}
    \label{equ:equ4}
    r_{\theta}(t,x):=\mathcal{N}[u_{\theta}(t,x)]-f(t,x).
\end{equation}
The weights $\theta$ are trained by minimizing a loss function:
\begin{equation}
    \label{equ:equ5}
    \mathcal{L}(\theta)=\mathcal{L}_{r}(\theta)+\mathcal{L}_{b}(\theta)+\mathcal{L}_{0}(\theta),
\end{equation}
where $\mathcal{L}_{r}$, $\mathcal{L}_{b}$, and $\mathcal{L}_{0}$ are loss terms that penalize the PDE \eqref{equ:equ1}, boundary condition \eqref{equ:equ2}, initial condition \eqref{equ:equ3}, respectively:
\begin{equation}
    \label{equ:equ6}
    \mathcal{L}_{r}(\theta)=\frac{1}{N_{r}}\sum\limits_{i=1}^{N_{r}}\vert r_{\theta}(t_{r}^{i},x_{r}^{i})\vert^{2},
\end{equation}
\begin{equation}
    \label{equ:equ7}
    \mathcal{L}_{b}(\theta)=\frac{1}{N_{b}}\sum\limits_{i=1}^{N_{b}}\vert\mathcal{B}[u_{\theta}(t_{b}^{i},x_{b}^{i})]-g(t_{b}^{i},x_{b}^{i})\vert^{2},
\end{equation}
\begin{equation}
    \label{equ:equ8}
    \mathcal{L}_{0}(\theta)=\frac{1}{N_{0}}\sum\limits_{i=1}^{N_{0}}\vert u_{\theta}(0,x_{0}^{i})-h(x_{0}^{i})\vert^{2},
\end{equation}
where $\{t_{r}^{i},x_{r}^{i}\}_{i=1}^{N_{r}}$ are collocation points randomly sampled from the domain $\Omega$, $\{t_{b}^{i},x_{b}^{i}\}_{i=1}^{N_{b}}$ are boundary condition points, $\{t_{0}^{i},x_{0}^{i}\}_{i=1}^{N_{0}}$ are initial condition points, and $N_{r}$, $N_{b}$, and $N_{0}$ denote the number of collocation, boundary, initial points, respectively. The structure of vanilla PINNs is shown in Figure \ref{fig:fig1}.

\subsection{Related work}
In the previous section, the most basic idea of vanilla PINNs was described. Although PINNs have attracted many attentions in recent years, handling high dimensional and complex PDE problems remains challenging. Further studies need to be done to achieve higher training efficiency, more accurate prediction, and faster convergence speed.
\subsubsection{Neural network architecture}
Many people \cite{lagari2020systematic,lu2021physics,dong2021method,demo2021extended,yang2022multi} modified neural network architectures to achieve a better result. In \cite{lagari2020systematic,lu2021physics,dong2021method}, boundary conditions are enforced more precisely by constructing special neural network architectures. Demo et~al. \cite{demo2021extended} aimed to divide the neural network into a combination of smaller neural networks, thus isolating the relations between the outputs of the model. Yang et~al.\cite{yang2022multi} used multiple neurons in the output layer so that knowledge regarding uncertainty can be imposed on the distribution formed by outputs.
\subsubsection{Domain decomposition}
Krishnapriyan et~al. \cite{krishnapriyan2021characterizing} proved that failure modes caused by soft regularization in PINNs can make the loss harder to optimize. They proposed two approaches to tackle these failure modes, one is to use curriculum regularization, where a good initialization is found to warm start the training, the other is to view the problem as a sequence-to-sequence(seq2seq) learning task to predict the solution at next time step, rather than learning the entire space-time, and this idea is also mentioned in \cite{wang2022respecting,wight2020solving,mattey2022novel}.
\subsubsection{Parameter initialization}
Xavier initialization \cite{glorot2010understanding} is widely used by researchers to select appropriate initial weights and biases, but few consider the influence of initialization. Liu et~al. \cite{liu2022novel} proposed a meta-learning initialization method based on labeled data, which allows PINNs to converge faster and predict more accurately.
\subsubsection{Loss balancing}
Different parts of the loss function tend to descend at different speeds using gradient descent, which restricts weights and biases to converge to the global optimal solution. So most of the further studies on PINNs focus on balancing different parts of loss function rather than treating them equally. The standard approach is to introduce weights into Equation \eqref{equ:equ5}:
\begin{equation}
    \label{equ:equ9}
    \mathcal{L}(\theta)=\lambda_{r}\mathcal{L}_{r}(\theta)+\lambda_{b}\mathcal{L}_{b}(\theta)+\lambda_{0}\mathcal{L}_{0}(\theta).
\end{equation}
Wight and Zhao \cite{wight2020solving} pointed out that it is significant to enforce neural networks to fit the initial conditions exactly, especially for equations that can only be solved in the forward time direction. Accordingly, the loss function was redefined as $\mathcal{L}(\theta)=\mathcal{L}_{r}(\theta)+\mathcal{L}_{b}(\theta)+C_{0}\mathcal{L}_{0}(\theta)$, where $C_{0}$ is a big positive constant that is tunable. In \cite{wang2021understanding}, the authors, drawing motivation from Adam \cite{kingma2014adam}, proposed a learning rate annealing procedure, which was designed to scale the loss by utilizing the back-propagated gradient statistics automatically. Yuan et~al. \cite{yuan2022pinn} adopted an adaptive strategy, where the weights of components are proportional to the values of residuals, that is the weight of the smallest residual component will be normalized to 1, and greater weights will be assigned to larger components. SoftAdapt \cite{heydari2019softadapt} balances the loss terms using the rate of change of each loss term between consecutive steps. Bischof and Kraus \cite{bischof2021multi} proposed a novel method taking into consideration the advantages of \cite{wang2021understanding} and \cite{heydari2019softadapt}, where gradient statistics were avoided and drastic changes in the loss space were reduced.
\subsubsection{Uncertainty quantification}
Quantifying uncertainty in PINNs is still a challenging problem. Yang et~al. \cite{yang2021b} proposed a Bayesian physics-informed neural network(B-PINN) to provide predictions and quantify the aleatoric uncertainty arising from the noisy data. In \cite{zhang2019quantifying}, the random parameter in the equation was represented as a stochastic process, introducing parametric uncertainty, dropout \cite{srivastava2014dropout} was also used to estimate the approximation uncertainty, the parametric uncertainty, and the approximation uncertainty were referred to as total uncertainty. Yang and Perdikaris \cite{yang2019adversarial} employed latent variable models to construct probabilistic representations so that uncertainty can be quantified and propagated.

Literature mentioned above all modeled the uncertainty in an implicit or indirect way, motivated by \cite{nix1994estimating,lakshminarayanan2017simple,pestourie2020active}, we output two values in the final layer of vanilla PINNs, representing the predicted mean and variance directly, and we also incorporate the variance into the training of PINNs, which speeds up the convergence speed.

\section{Methodology}
\label{sec:methodology}
The goal of this paper is to take into consideration the influence of uncertainty during the process of training. In this section, we will give a detailed description of the proposed VI-PINNs.
\subsection{Architecture of VI-PINNs}
Vanilla PINNs put together the PDE, boundary conditions, and initial conditions into the total loss function, and the weights and biases are trained using certain optimization algorithms.

\begin{figure}[htbp]
    \centering
    \includegraphics[scale=0.15]{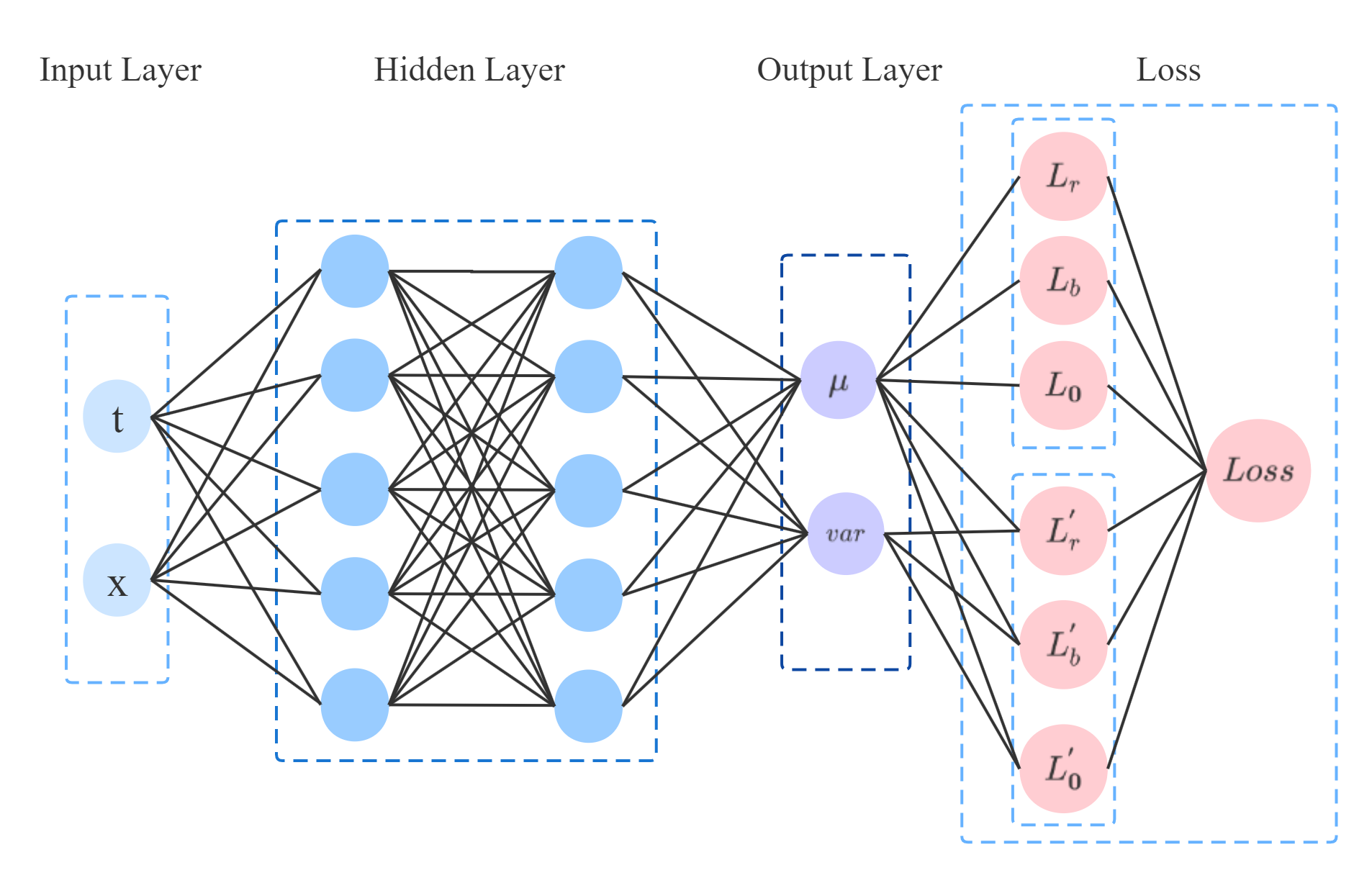}
    \caption{Schematic of VI-PINNs}
    \label{fig:fig2}
\end{figure}

VI-PINNs is an extension of vanilla PINNs. We want to represent the model uncertainty directly, so we use a heteroscedastic neural network to output the predicted mean $\mu(x)$ and the variance $\sigma^{2}(x)$ in the final layer. Notably, in order to enforce variance to obey the positivity constraint, following \cite{lakshminarayanan2017simple}, we pass the second output $\sigma^{2}(x)$ through the \emph{softplus} function $\log(1+\exp(\cdot))$, and add a very small constant $10^{-6}$ to the variance for numerical stability. The structure of VI-PINNs is shown in Figure \ref{fig:fig2}.

Having decided the particular network architecture, how to combine the predicted mean $\mu(x)$ and the variance $\sigma^{2}(x)$ needs to be addressed. Naturally, we can minimize the negative log-likelihood criterion by considering a Gaussian assumption on target distribution, as described in Section \ref{sub:sub32}. Mean square error and negative log-likelihood both have their advantages on regression problems, therefore we would like to incorporate them. Considering the successful application of mean square error in PINNs, we figure out a solution that negative log-likelihood can be used to assist the training procedure of mean square error, which is the auxiliary task mentioned in Section \ref{sub:sub33}.
\subsection{NLL loss for VI-PINNs}
\label{sub:sub32}
For regression problems, mean square error (MSE) is always taken as the loss function for optimizing. However, MSE alone does not have the ability to capture uncertainty. In order to capture uncertainty, we view the observed value as a sample from a Gaussian distribution, so we have:
\begin{equation}
    \label{equ:equ10}
    P(y_{i}\vert x_{i})=\frac{1}{\sqrt{2\pi\sigma_{\theta}^{2}(x_{i})}}\exp\left(\frac{-[y_{i}-\mu_{\theta}(x_{i})]^{2}}{2\sigma_{\theta}^{2}(x_{i})}\right).
\end{equation}
If we take the natural log of both sides, we will get the following form:
\begin{equation}
    \label{equ:equ11}
    -\log P(y_{i}\vert x_{i})= \frac{\log [\sigma_{\theta}^{2}(x_{i})]}{2} +  \frac{[y_{i}-\mu_{\theta}(x_{i})]^{2}}{2\sigma_{\theta}^{2}(x_{i})} + \frac{\log (2\pi)}{2}.
\end{equation}
Accordingly, negative log-likelihood loss (NLL) is taken for better optimizing model parameters over all $x_{i}$:
\begin{equation}
\label{equ:equ12}
    \mathcal{L}_{NLL} = \frac{1}{N}\sum \limits_{i=1}^{N} \left(\frac{\log[\sigma^{2}_{\theta}(x_{i})]}{2}+\frac{[y_{i}-\mu_{\theta}(x_{i})]^{2}}{2\sigma^{2}_{\theta}(x_{i})}+\frac{\log (2\pi)}{2}\right).
\end{equation}
The last term is a constant and can be ignored for optimization, the simplified form of Equation \eqref{equ:equ12} is:
\begin{equation}
\label{equ:equ13}
    \mathcal{L}_{NLL} = \frac{1}{N}\sum \limits_{i=1}^{N} \left(\frac{\log[\sigma^{2}_{\theta}(x_{i})]}{2}+\frac{[y_{i}-\mu_{\theta}(x_{i})]^{2}}{2\sigma^{2}_{\theta}(x_{i})}\right).
\end{equation}
However, the NLL loss above does not perform satisfactorily well in our experiments. In our opinion, as the training process goes on, the variance $\sigma^{2}(x)$ becomes smaller and smaller, $\log\sigma^{2}(x)$ becomes a big negative value, which might not be helpful to the training. So we adjust the NLL loss slightly, and the $\log\sigma^{2}(x)$ will not be a negative value even if $\sigma^{2}(x)<<1$, the modified negative log-likelihood(MNLL) loss is represented as the following form:
\begin{equation}
\label{equ:equ14}
    \mathcal{L}_{MNLL} = \frac{1}{N} \sum \limits_{i=1}^{N}\left(\log[\sigma^{2}_{\theta}(x)+1]+\frac{[y-\mu_{\theta}(x)]^{2}}{\sigma^{2}_{\theta}(x)}\right).
\end{equation}
Applying the loss function to Equation \eqref{equ:equ1}-\eqref{equ:equ3}, we have the following form of $\mathcal{L}_{r}^{'}(\theta)$, $\mathcal{L}_{b}^{'}(\theta)$, and $\mathcal{L}_{0}^{'}(\theta)$ accordingly:
\begin{equation}
    \label{equ:equ15}
    \mathcal{L}_{r}^{'}(\theta)=\frac{1}{N_{r}}\sum\limits_{i=1}^{N_{r}}\left(\log[ \sigma^{2}_{\theta}(t_{r}^{i},x_{r}^{i})+1]+\frac{[\mathcal{N}[u_{\theta}(t_{r}^{i},x_{r}^{i})]-f(t_{r}^{i},x_{r}^{i})]^{2}}{\sigma^{2}_{\theta}(t_{r}^{i},x_{r}^{i})}\right),
\end{equation}
\begin{equation}
    \label{equ:equ16}
    \mathcal{L}_{b}^{'}(\theta)=\frac{1}{N_{b}}\sum\limits_{i=1}^{N_{b}}\left(\log[ \sigma^{2}_{\theta}(t_{b}^{i},x_{b}^{i})+1]+\frac{[\mathcal{B}[u_{\theta}(t_{b}^{i},x_{b}^{i})]-g(t_{b}^{i},x_{b}^{i})]^{2}}{\sigma^{2}_{\theta}(t_{b}^{i},x_{b}^{i})}\right),
\end{equation}
\begin{equation}
    \label{equ:equ17}
    \mathcal{L}_{0}^{'}(\theta)=\frac{1}{N_{0}}\sum\limits_{i=1}^{N_{0}}\left(\log[ \sigma^{2}_{\theta}(t_{0}^{i},x_{0}^{i})+1]+\frac{[u_{\theta}(0,x_{0}^{i})-h(x_{0}^{i})]^{2}}{\sigma^{2}_{\theta}(t_{0}^{i},x_{0}^{i})}\right),
\end{equation}
The loss $\mathcal{L}^{'}(\theta)$ is composed by loss term related to variance $\mathcal{L}^{'}_{r}(\theta)$, $\mathcal{L}^{'}_{b}(\theta)$ and $\mathcal{L}^{'}_{0}(\theta)$:
\begin{equation}
    \label{equ:equ18}
    \mathcal{L}^{'}(\theta)=\mathcal{L}_{r}^{'}(\theta)+\mathcal{L}_{b}^{'}(\theta)+\mathcal{L}_{0}^{'}(\theta).
\end{equation}
\subsection{Auxiliary task in VI-PINNs}
\label{sub:sub33}
All machine learning tasks have certain goals. By introducing some related tasks with similar goals, the model can learn more useful information to improve the performance of the tasks we are concerned about, and these related tasks are called auxiliary tasks. Taking \cite{liebel2018auxiliary} as an example, the authors chose SIDE and semantic segmentation as the main tasks. By adding auxiliary tasks such as regression for the current time of day and classification of weather conditions, the model is able to learn a rich and robust common representation of an image, resulting in improved performance on the main tasks. It has been shown that auxiliary tasks can improve the performance, robustness, and training speed of networks.

Instead of using the $\mathcal{L}^{'}(\theta)$ solely to obtain the optimal model parameters, we prefer to introduce it to the vanilla PINNs loss function as an auxiliary task. For one thing, existing studies on PINNs have proved that the loss function proposed in \cite{raissi2019physics} is simple but sufficient to represent what we want to learn from data to some extent, there is no need to utilize a totally novel loss function; for another, \cite{jaderberg2016reinforcement, liebel2018auxiliary, ye2020auxiliary, lyle2021effect} conveyed to us that by introducing auxiliary tasks as complementary knowledge, the performance of the main task is expected to boost.

So we take the following loss function as the overall loss function of our VI-PINNs:
\begin{equation}
    \label{equ:equ19}
    \mathcal{L}_{total} = \mathcal{L}(\theta)+\lambda\mathcal{L}^{'}(\theta),
\end{equation}
where $\mathcal{L}(\theta)$ and $\mathcal{L}^{'}(\theta)$ are defined in detail in Equation \eqref{equ:equ5} and \eqref{equ:equ18} respectively, the weight $\lambda$ is a hyperparameter needs to be adjusted manually to better assist the training process under different weights and bias initialization.

\section{Results}
\label{sec:results}
In this section, we provide some numerical examples to demonstrate the performance of our proposed VI-PINNs against some other methods. The methods we compared are summarized in Table \ref{table:table1}.

\begin{table}[htbp]
\renewcommand{\arraystretch}{1.2}
\begin{center}
\begin{minipage}{300pt}
\caption{Methods considered in the numerical examples}
\label{table:table1}
\begin{tabular}{@{}ll@{}}
\toprule
Method & Detail    \\                                                                           \midrule
M1     & Vanilla PINN proposed in \cite{raissi2019physics} \\
M2     & \begin{tabular}[l]{@{}l@{}} Multiplying the initial condition term of vanilla loss function \\ by a big positive constant \cite{wight2020solving} \end{tabular} \\
M3     & \begin{tabular}[l]{@{}l@{}} Adjusting the weights of different terms of vanilla loss function \\ by normalizing the values of different terms \cite{yuan2022pinn}  \end{tabular} \\
M4     & \begin{tabular}[l]{@{}l@{}}Adjusting the weights of different terms of vanilla loss function \\ using SoftAdapt \cite{heydari2019softadapt}\end{tabular} \\
M5     & Our proposed model considering auxiliary task \\
\bottomrule
\end{tabular}
\end{minipage}
\end{center}
\end{table}

In all examples, the activation function is $tanh(\cdot)$, a differentiable nonlinear activation function. The optimizer is Adam \cite{kingma2014adam} with a fixed learning rate 0.001. All codes are implemented in Pytorch \cite{paszke2019pytorch}, a framework used widely in machine learning. In all cases below, we use fully connected networks containing three hidden layers, each with 30 neurons. In our experiments, we use full batch training rather than mini-batch training for our architecture is very simple. We train networks on a small dataset, and we test the performance on a larger dataset as the more points we use for testing, the more accurate the results will be. Mean square error (MSE) and $L_{2}$-error are used as the metric for testing the performance of the trained model.
\begin{equation}
    MSE=\frac{1}{N}\sum\limits_{i=1}^{N}\vert u(t_{i},x_{i})-\hat{u}(t_{i},x_{i})\vert^{2},
    \label{equ:equ20}
\end{equation}
\begin{equation}
    L_{2}{\rm \mbox{-} error}=\frac{\sqrt{\sum\limits_{i=1}^{N}\vert u(t_{i},x_{i})-\hat{u}(t_{i},x_{i})\vert^{2}}}{\sqrt{\sum\limits_{i=1}^{N}\vert u(t_{i},x_{i})\vert^{2}}},
    \label{equ:equ21}
\end{equation}
where $u$ is the reference solution, $\hat{u}$ is the predicted solution.

To get rid of the effect of randomness in PINNs, we run the same experiment five times on the same dataset but with different random seeds of initialization to ensure that the results can be reproduced, and the mean errors of these five times are used to represent final errors. In the last subsection below, we also perform simple systematic studies of a simple equation to quantify predictive accuracy for different choices of the number of points, different network architectures, and hyperparameter $\lambda$.

\subsection{Advection Equation}
\label{sub:sub41}
The first example we consider is the advection equation,
\begin{equation}
    \label{equ:equ22}
    \frac{\partial u}{\partial t}+\frac{\partial u}{\partial x}=0, \ x\in[0,1], \ t\in[0,0.5],
\end{equation}
with the initial condition,
\begin{equation}
    \label{equ:equ23}
    u(0,x)=2sin(\pi x),
\end{equation}
and boundary conditions,
\begin{equation}
    \label{equ:equ24}
    u(t,0)=-2sin(\pi t), \ u(t,1)=2sin(\pi t),
\end{equation}
where the exact solution for this advection equation is,
\begin{equation}
    \label{equ:equ25}
    u(t,x)=2sin(\pi(x-t)).
\end{equation}

\begin{figure}[htbp]
    \centering
    \includegraphics[width=1\textwidth]{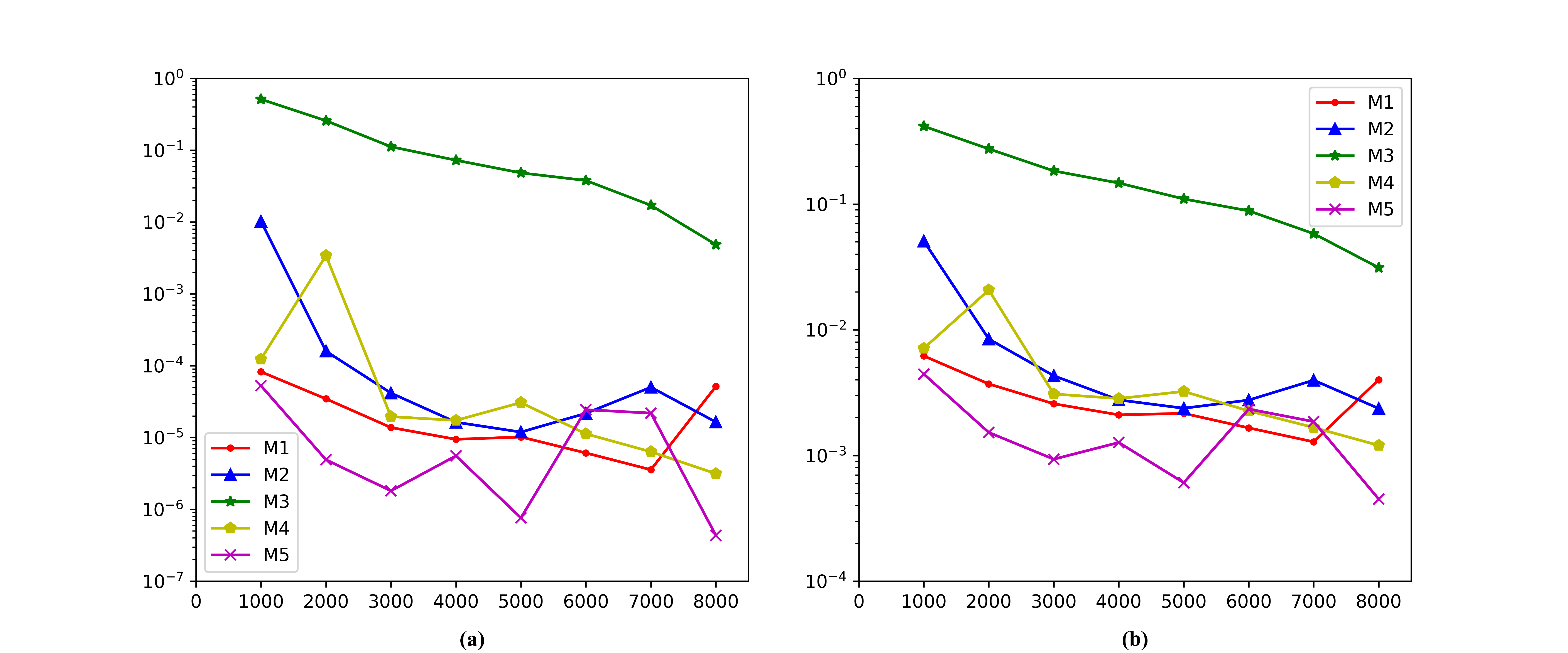}
    \caption{MSE and $L_{2}$-error values in iteration for the advection equation \textbf{(a)} MSE for the advection equation \textbf{(b)} $L_{2}$-error for the advection equation}
    \label{fig:fig3}
\end{figure}

Compared to the other equations, the advection equation is easy to fit, so the total number of iterations is set to 8,000, and the number of collocation points, initial points, and boundary points are set to 500, 100, and 100, respectively. The performance of different methods is evaluated on a large dataset with 5,000 points. Table \ref{table:table2} displays MSE and $L_{2}$-error of all methods and Figure \ref{fig:fig3} represents the MSE and $L_{2}$-error values in iteration.

\begin{table}[htbp]
\renewcommand{\arraystretch}{1.2}
\begin{center}
\begin{minipage}{300pt}
\caption{Statistics of MSE and $L_2$-error for the advection equation}
\label{table:table2}
\begin{tabular}{lccccc}
\toprule
                   & M1       & M2       & M3       & M4       & M5                \\
\midrule
Mean MSE           & 2.62E-06 & 5.26E-06 & 4.86E-03 & 3.14E-06 & \textbf{4.31E-07} \\
Mean $L_{2}$-error & 1.09E-03 & 1.58E-03 & 3.12E-02 & 1.20E-03 & \textbf{4.48E-04} \\
Min MSE            & 1.06E-06 & 2.92E-06 & 4.39E-05 & 1.12E-06 & \textbf{1.94E-07} \\
Min $L_{2}$-error  & 7.24E-04 & 1.20E-03 & 4.66E-03 & 7.43E-04 & \textbf{3.10E-04} \\
Max MSE            & 5.44E-06 & 8.86E-06 & 2.23E-02 & 5.44E-06 & \textbf{7.70E-07} \\
Max $L_{2}$-error  & 1.64E-03 & 2.09E-03 & 1.05E-01 & 1.64E-03 & \textbf{6.17E-04} \\
\bottomrule
\end{tabular}
\end{minipage}
\end{center}
\end{table}

As shown in Table \ref{table:table2}, our proposed method gets the best result within five methods, even the worst performance can achieve an $L_{2}$-error of 6.17e-04, while results of other methods are all bigger than 1e-03. As shown in Figure \ref{fig:fig3}, our proposed method can get more accurate results while taking fewer training steps, while other methods get results that are close to each other except M3.

\subsection{Burgers Equation}
The Burgers equation considered here is defined as,
\begin{equation}
    \label{equ:equ26}
    \frac{\partial u}{\partial t}+u\frac{\partial u}{\partial x}-\frac{0.01}{\pi}\frac{\partial^{2}u}{\partial x^{2}}=0, \ x\in[-1,1],\ t\in[0,1],
\end{equation}
\begin{equation}
    \label{equ:equ27}
    u(0,x)=-sin(\pi x),
\end{equation}
\begin{equation}
    \label{equ:equ28}
    u(t,-1)=u(t,1)=0.
\end{equation}

\begin{figure}[htbp]
    \centering
    \includegraphics[width=1\textwidth]{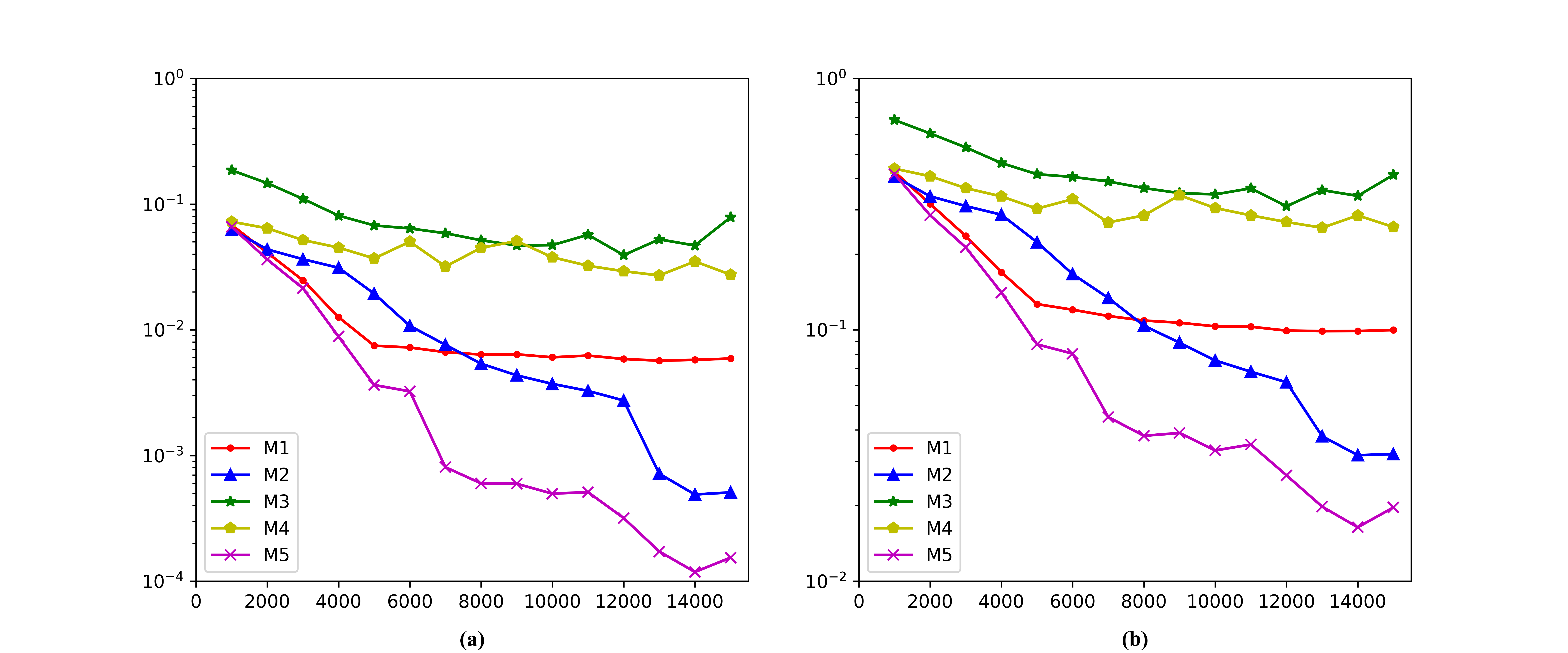}
    \caption{MSE and $L_{2}$-error values in iteration for the burgers equation \textbf{(a)} MSE for the burgers equation \textbf{(b)} $L_{2}$-error for the burgers equation}
    \label{fig:fig4}
\end{figure}

\begin{figure}[H]
    \centering
    \includegraphics[scale=0.25]{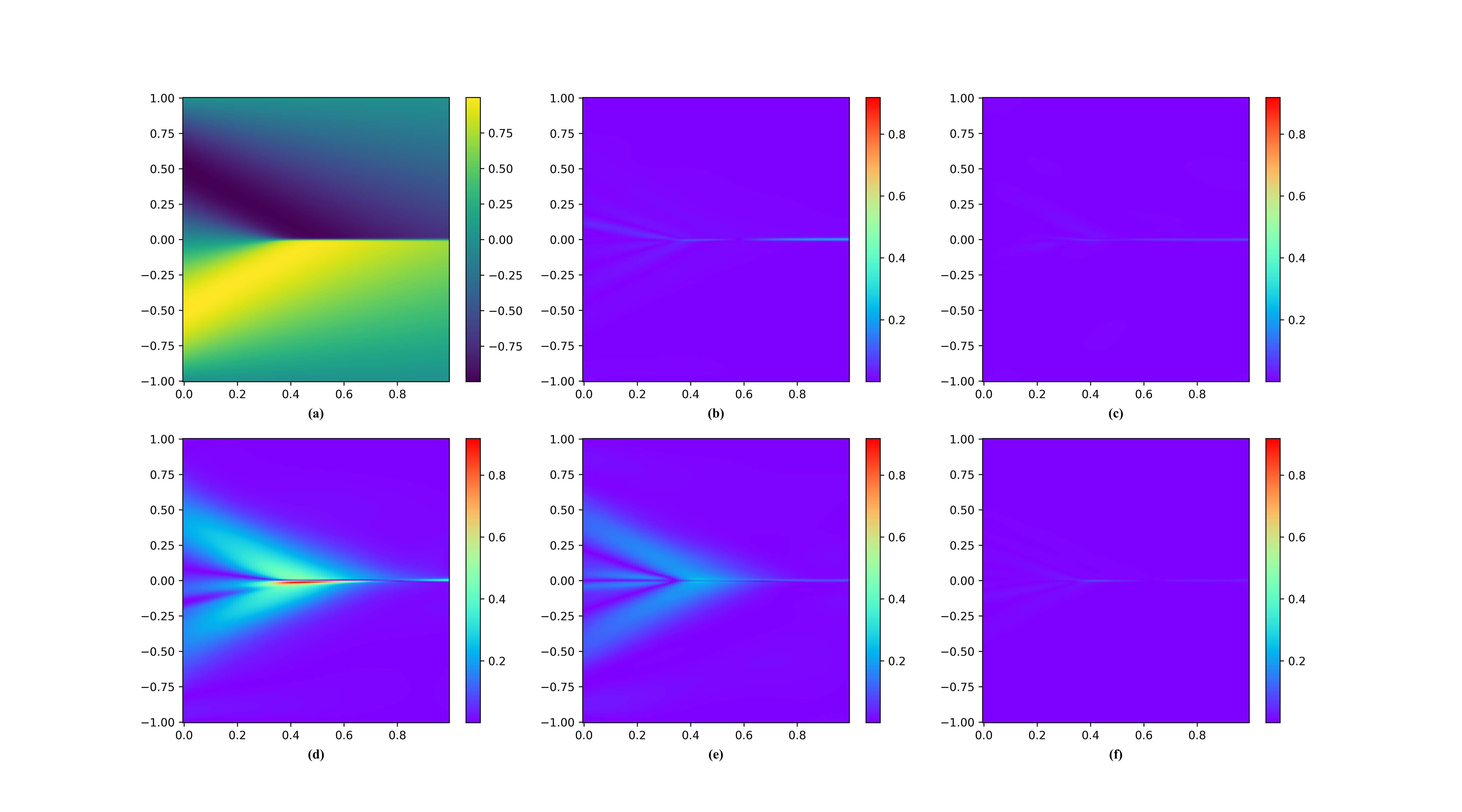}
    \caption{The exact solution and the best absolute error of different methods for the burgers equation \textbf{(a)} The exact solution of the burgers equation \textbf{(b)-(f)} The best absolute error of different methods}
    \label{fig:fig5}
\end{figure}

Unlike other equations having a smooth solution, the solution of the burgers equation has a sharp internal layer. The total number of iterations is set to 15,000 and the number of collocation points, initial points, and boundary points is set to 5000, 500, and 500, respectively. We test the performance of different methods on 25,600 points used in \cite{raissi2019physics}. Table \ref{table:table3} displays MSE and $L_{2}$-error of all methods and Figure \ref{fig:fig4} represents the MSE and $L_{2}$-error values in iteration. Figure \ref{fig:fig5} represents the exact solution and the best absolute error of different methods for the burgers equation.

As shown in Table \ref{table:table3}, our proposed method gets the best performance, M2 also performs well, that is to say, for the burgers equation, emphasizing the effect of initial condition points helps to improve the performance as well. From Figure \ref{fig:fig4} we can conclude that as the training process goes on, our method converges faster than other methods, and M1, M3, and M4 hardly decline later. The solution of the burgers equation has a very steep domain when $x=0$. Figure \ref{fig:fig5} shows that M1, M2, and M5 are all able to capture this feature, but our result fits the feature more closely compared to the other methods.

\begin{table}[htbp]
\renewcommand{\arraystretch}{1.2}
\begin{center}
\begin{minipage}{300pt}
\caption{Statistics of MSE and $L_2$-error for the burgers equation}
\label{table:table3}
\begin{tabular}{lccccc}
\toprule
                   & M1       & M2       & M3       & M4       & M5                \\
\midrule
Mean MSE           & 5.44E-03 & 4.55E-04 & 2.67E-02 & 1.40E-02 & \textbf{1.05E-04} \\
Mean $L_{2}$-error & 9.46E-02 & 3.02E-02 & 2.61E-01 & 1.75E-01 & \textbf{1.46E-02} \\
Min MSE            & 1.26E-04 & 2.44E-05 & 1.38E-02 & 2.84E-03 & \textbf{1.06E-05} \\
Min $L_{2}$-error  & 1.83E-02 & 8.05E-03 & 1.91E-01 & 8.67E-02 & \textbf{5.29E-03} \\
Max MSE            & 1.35E-02 & 8.72E-04 & 4.07E-02 & 3.79E-02 & \textbf{2.83E-04} \\
Max $L_{2}$-error  & 1.89E-01 & 4.81E-02 & 3.29E-01 & 3.17E-01 & \textbf{2.74E-02} \\
\bottomrule
\end{tabular}
\end{minipage}
\end{center}
\end{table}

\subsection{Convection-Diffusion Equation}
\begin{figure}[htbp]
    \centering
    \includegraphics[width=1\textwidth]{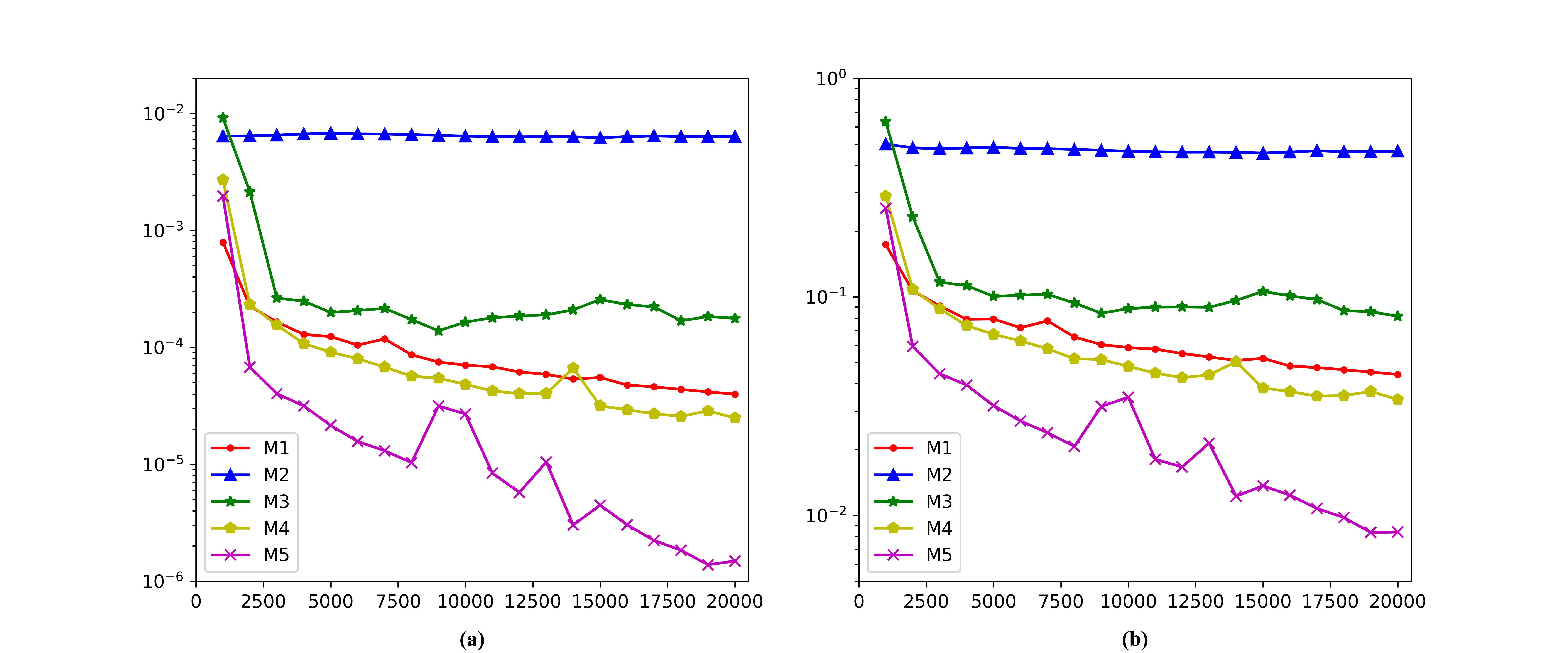}
    \caption{MSE and $L_{2}$-error values in iteration for the convection-diffusion equation \textbf{(a)} MSE for the convection-diffusion equation \textbf{(b)} $L_{2}$-error for the convection-diffusion equation}
    \label{fig:fig6}
\end{figure}

\begin{figure}[H]
    \centering
    \includegraphics[scale=0.25]{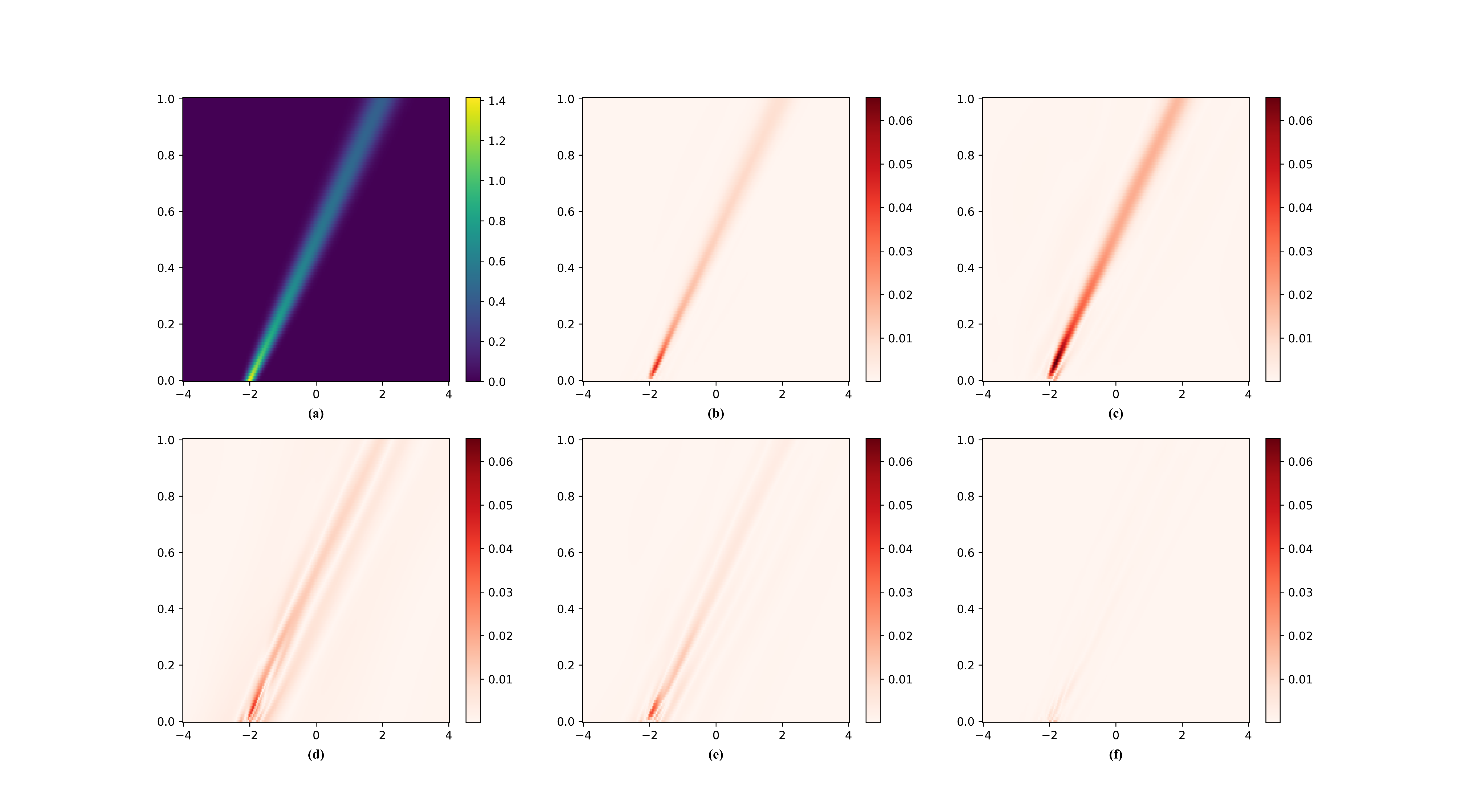}
    \caption{The exact solution and the best absolute error of different methods for the convection-diffusion equation \textbf{(a)} The exact solution of the convection-diffusion equation \textbf{(b)-(f)} The best absolute error of different methods}
    \label{fig:fig7}
\end{figure}

Convection-diffusion equation is a linear PDE, containing an advection term and a diffusion term. The equation is defined as follows:
\begin{equation}
    \label{equ:equ29}
    \frac{\partial u}{\partial t}+c \cdot \frac{\partial u}{\partial x}=\mu \cdot \frac{\partial^{2}u}{\partial x^{2}}, \ x\in[-L,L], \ t\in[0,T],
\end{equation}
\begin{equation}
    \label{equ:equ30}
    u(0,x)=\frac{0.1}{\sqrt{0.1\cdot \mu}}\cdot\exp\left(-\frac{(x+2)^{2}}{4\cdot 0.1 \cdot \mu}\right),
\end{equation}
\begin{equation}
    \label{equ:equ31}
    u(t,-L)=u(t,L)=0,
\end{equation}
where the exact solution is:
\begin{equation}
    \label{equ:equ32}
    u(t,x)=\frac{0.1}{\sqrt{(t+0.1)\cdot\mu}}\exp\left(-\frac{(x+2-4t)^{2}}{4\cdot(t+0.1)\cdot\mu}\right)
\end{equation}
Here we take $c=4.0$, $\mu=0.05$, $L=4$, $T=1$ adopted in \cite{peng2022rang}, the total number of iterations is set to 20,000 and the number of collocation points, initial points, and boundary points are set to 1,000, 100, and 100, respectively. The performance of different methods is evaluated on a large dataset with 5,000 points. Table \ref{table:table4} displays MSE and $L_{2}$-error of all methods and Figure \ref{fig:fig6} represents the MSE and $L_{2}$-error values in iteration. Figure \ref{fig:fig7} depicts the exact solution and the best absolute error between the exact solution and predicted solutions of the convection-diffusion equation.

\begin{table}[htbp]
\renewcommand{\arraystretch}{1.2}
\begin{center}
\begin{minipage}{300pt}
\caption{Statistics of MSE and $L_2$-error for the convection-diffusion equation}
\label{table:table4}
\begin{tabular}{lccccc}
\toprule
                   & M1       & M2       & M3       & M4       & M5                \\
\midrule
Mean MSE           & 3.84E-05 & 5.41E-03 & 8.86E-05 & 2.32E-05 & \textbf{1.10E-06} \\
Mean $L_{2}$-error & 4.32E-02 & 4.25E-01 & 6.22E-02 & 3.24E-02 & \textbf{7.25E-03} \\
Min MSE            & 7.95E-06 & 2.90E-05 & 1.12E-05 & 3.85E-06 & \textbf{1.91E-07} \\
Min $L_{2}$-error  & 2.07E-02 & 3.96E-02 & 2.46E-02 & 1.44E-02 & \textbf{3.21E-03} \\
Max MSE            & 6.93E-05 & 1.51E-02 & 2.39E-04 & 5.36E-05 & \textbf{2.01E-06} \\
Max $L_{2}$-error  & 6.12E-02 & 9.03E-01 & 1.14E-01 & 5.38E-02 & \textbf{1.04E-02} \\
\bottomrule
\end{tabular}
\end{minipage}
\end{center}
\end{table}

As shown in Table \ref{table:table4}, within these five methods, our proposed method significantly outperforms the others. The MSE and $L_{2}$-error of M2 hardly decline, meaning that in this equation putting emphasis on the effect of initial conditions may wind up with the opposite effect. M5 gets a better result compared with M1, denoting that it is practical to adopt a multi-object optimization strategy. From Figure \ref{fig:fig7}, we can get that our method is capable of giving a more exact solution over the whole domain in contrast to the remaining methods.

\subsection{Poisson Equation}
\begin{figure}[htbp]
    \centering
    \includegraphics[scale=0.3]{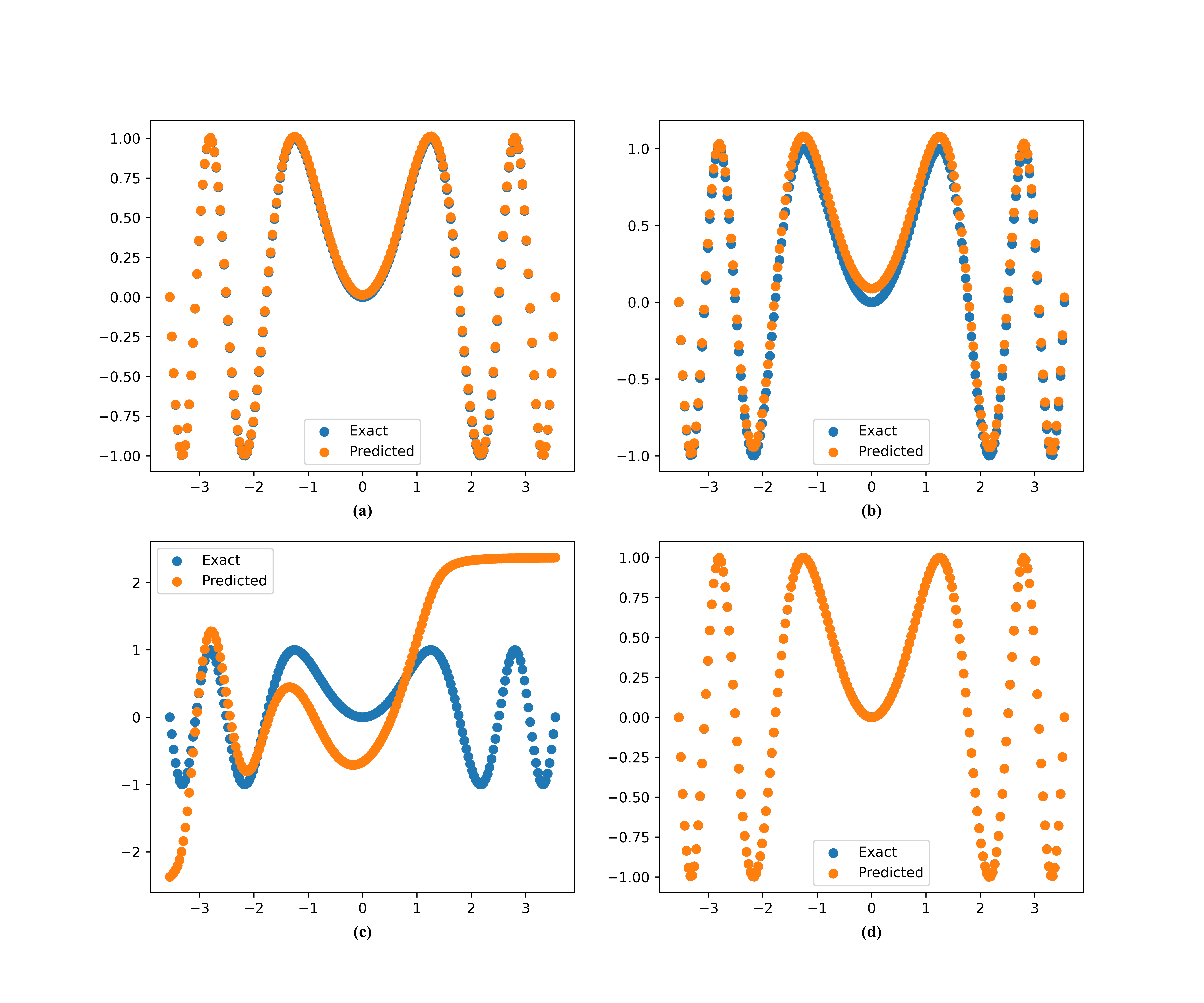}
    \caption{Contrast between the exact and the predicted solutions for the Poisson equation}
    \label{fig:fig8}
\end{figure}

The general form of the poisson equation is:
\begin{equation}
    \label{equ:equ33}
    -u'' = f(x), \ x\in[-L,L],
\end{equation}
with boundary conditions:
\begin{equation}
    \label{equ:equ34}
    u(-L)=u(L)=0.
\end{equation}
Here, we take $L=2\sqrt{\pi}$ and consider using a special exact solution $u(x)=sin(x^{2})$, as x moves away from 0, the solution becomes oscillating. Now, we can fabricate the $f(x)$ by derivation, the specific form becomes:
\begin{equation}
    \label{equ:equ35}
    -u'' = 4x^{2}sin(x^{2})-2cos(x^{2}), \ x\in[-2\sqrt{\pi},2\sqrt{\pi}],
\end{equation}
\begin{equation}
    \label{equ:equ36}
    u(-2\sqrt{\pi})=u(2\sqrt{\pi})=0.
\end{equation}
The total number of iterations is set to 10,000, 2,500 points are sampled randomly for collocation points and we just need two points for boundary conditions. The performance of different methods is evaluated on a large dataset with 5,000 points. We display the mean MSE and $L_{2}$-error of five duplicate experiments in Table \ref{table:table5}, and we also represent the contrast between the exact and the best predicted solutions in Figure \ref{fig:fig8}.

\begin{table}[htbp]
\renewcommand{\arraystretch}{1.2}
\begin{center}
\begin{minipage}{250pt}
\caption{Statistics of MSE and $L_2$-error for the Poisson equation}
\label{table:table5}
\begin{tabular}{lcccc}
\toprule
                   & M1       & M2       & M3       & M5                \\
\midrule
Mean MSE           & 2.10E-05 & 3.88E-03 & 4.06E+00 & \textbf{2.89E-07} \\
Mean $L_{2}$-error & 4.59E-03 & 8.56E-02 & 2.86E+00 & \textbf{7.43E-04} \\
Min MSE            & 8.14E-07 & 3.48E-04 & 8.15E-01 & \textbf{1.05E-07} \\
Min $L_{2}$-error  & 1.35E-03 & 2.80E-02 & 1.35E+00 & \textbf{4.86E-04} \\
Max MSE            & 9.75E-05 & 8.14E-03 & 7.27E+00 & \textbf{7.97E-07} \\
Max $L_{2}$-error  & 1.48E-02 & 1.35E-01 & 4.04E+00 & \textbf{1.34E-03} \\
\bottomrule
\end{tabular}
\end{minipage}
\end{center}
\end{table}

We do not take into consideration M4 in this equation since M4 balances the loss by utilizing the relative progress of the current time-step $t$ and the previous time-step $t-1$, which is practical theoretically but may exceed the computational range of computers for the loss of this equation may change drastically between two consecutive training epochs. As shown in Table \ref{table:table5}, M3 achieves the largest error given the number of training epochs for which it fails to fit the solution, as demonstrated in Figure \ref{fig:fig8}. Figure \ref{fig:fig8} depicts the exact and the predicted solutions for the poisson equation, our proposed method achieves the most accurate result among all the methods, we can also figure out that the neural networks approximate the solutions gradually, and our method converges faster than other methods.

\subsection{Wave Equation}
\begin{figure}[htbp]
    \centering
    \includegraphics[width=1\textwidth]{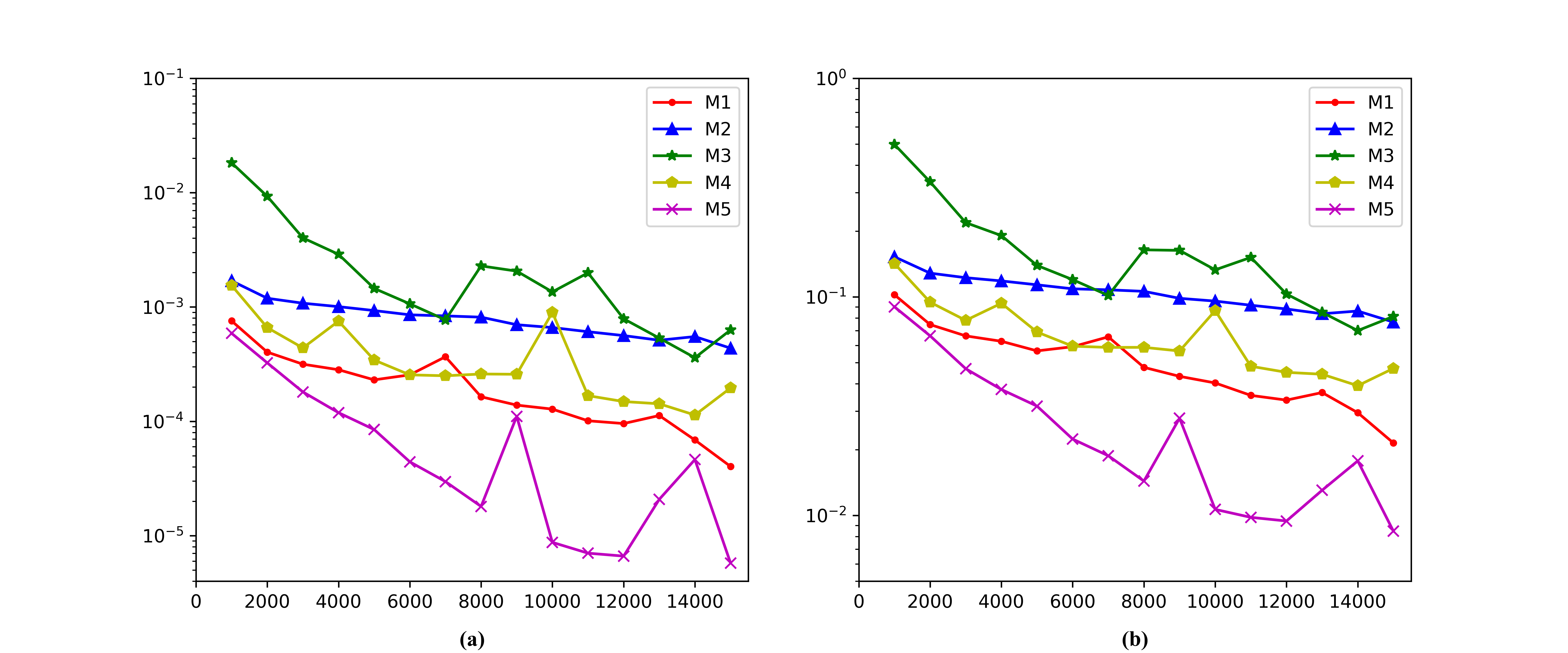}
    \caption{MSE and $L_{2}$-error values in iteration for the wave equation \textbf{(a)} MSE for the wave equation \textbf{(b)} $L_{2}$-error for the wave equation}
    \label{fig:fig9}
\end{figure}

\begin{figure}[H]
    \centering
    \includegraphics[scale=0.25]{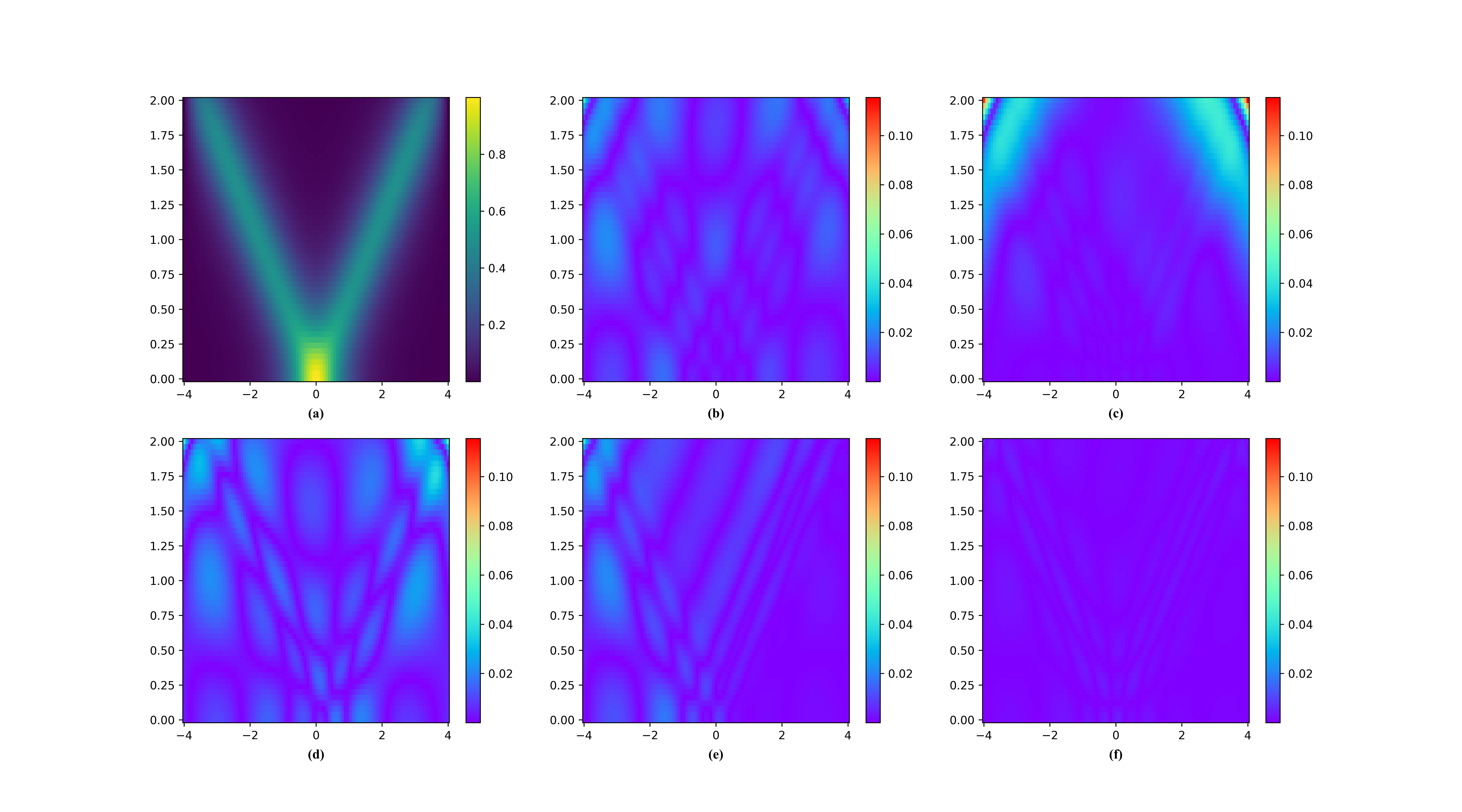}
    \caption{The exact solution and the best absolute error of different methods for the wave equation \textbf{(a)} The exact solution of the wave equation \textbf{(b)-(f)} The best absolute error of different methods}
    \label{fig:fig10}
\end{figure}

The wave equation to be discussed is defined as:
\begin{equation}
    \label{equ:equ37}
    \frac{\partial^{2}u}{\partial t^{2}}-3\frac{\partial^{2}u}{\partial x^{2}}=0, \ t \in [0,T], \ x \in [-L,L],
\end{equation}
with initial conditions:
\begin{equation}
    \label{equ:equ38}
    u(0,x)=\frac{1}{cosh(2x)}-\frac{0.5}{cosh(2(x-2L))}-\frac{0.5}{cosh(2(x+2L))},
\end{equation}
\begin{equation}
    \label{equ:equ39}
    \frac{\partial u}{\partial t}(0,x)=0,
\end{equation}
boundary conditions:
\begin{equation}
    \label{equ:equ40}
    u(t,-L)=u(t,L)=0,
\end{equation}
and the exact solution of this wave equation is:
\begin{equation}
\begin{split}
\label{equ:equ41}
     u(t,x)=\frac{0.5}{cosh(2(x-\sqrt{3}t))}-\frac{0.5}{cosh(2(x-2L+\sqrt{3}t))}\\
     +\frac{0.5}{cosh(2(x+\sqrt{3}t))}-\frac{0.5}{cosh(2(x+2L-\sqrt{3}t))}.
\end{split}
\end{equation}
Here, we take $T=2.0$, $L=4.0$, so $T$ and $L$ in Equation \eqref{equ:equ37}-\eqref{equ:equ41} can be substituted. The total number of iterations is set to 15,000 and the number of collocation points, initial points, partial initial points, and boundary points is set to 2,000, 200, 200, and 200, respectively. The performance of different methods is evaluated on a larger dataset with 5,000 points. Table \ref{table:table6} displays MSE and $L_{2}$-error of all methods and Figure \ref{fig:fig9} represents the MSE and $L_{2}$-error values in iteration. Figure \ref{fig:fig10} depicts the exact solution of the wave equation and the best absolute error between the exact solution and predicted solutions.

\begin{table}[htbp]
\renewcommand{\arraystretch}{1.2}
\begin{center}
\begin{minipage}{300pt}
\caption{Statistics of MSE and $L_2$-error for the wave equation}
\label{table:table6}
\begin{tabular}{lccccc}
\toprule
                   & M1       & M2       & M3       & M4       & M5                \\
\midrule
Mean MSE           & 4.03E-05 & 4.36E-04 & 2.99E-04 & 1.05E-04 & \textbf{4.15E-06} \\
Mean $L_{2}$-error & 2.15E-02 & 7.67E-02 & 6.32E-02 & 3.73E-02 & \textbf{7.45E-03} \\
Min MSE            & 3.69E-06 & 1.77E-04 & 1.14E-04 & 4.26E-05 & \textbf{1.72E-06} \\
Min $L_{2}$-error  & 7.17E-03 & 4.97E-02 & 3.99E-02 & 2.43E-02 & \textbf{4.89E-03} \\
Max MSE            & 7.01E-05 & 5.53E-04 & 4.71E-04 & 1.65E-04 & \textbf{6.15E-06} \\
Max $L_{2}$-error  & 3.12E-02 & 8.77E-02 & 8.09E-02 & 4.80E-02 & \textbf{9.25E-03} \\
\bottomrule
\end{tabular}
\end{minipage}
\end{center}
\end{table}

As shown in Figure \ref{fig:fig9}, the methods achieve similar results at the early training epochs. As the training process goes on, the result of our method drops significantly, while the others drop slightly. In this example, M2, M3, and M4 perform similarly. Figure \ref{fig:fig10} represents plainly the absolute error, our method achieves the smallest absolute error while other methods get larger errors across a large region of the domain.
\subsection{Comparative studies}
In this subsection, we perform a comparative study on the advection equation used in section \ref{sub:sub41} in order to figure out whether the choices of points, different network architectures, and hyperparameter $\lambda$ have big influences on the performance.

Table \ref{table:table7} summarizes the relationship between the point selection of initial points, boundary points $N_{u}$ and collocation points $N_{f}$, while the architecture is 3 hidden layers with 30 neurons per layer, the $\lambda$ is fixed to 1. By analyzing the table, we can deduce a general trend that given a fixed amount of collocation points $N_{f}$, the prediction becomes more accurate as the number of initial and boundary points $N_{u}$ increases, and vice versa. This is easy to explain, as the number of points increases, the model is able to acquire more useful knowledge, leading to improved performance.

\begin{table}[htbp]
\renewcommand{\arraystretch}{1.2}
\begin{center}
\begin{minipage}{190pt}
\caption{$L_{2}$-error under different choices of points for the advection equation}
\label{table:table7}
\begin{tabular}{cccc}
\toprule
\diagbox{$N_{u}$}{$N_{f}$}    & 200      & 500      & 1000     \\
\midrule
20  & 7.54E-04 & 6.20E-04 & 5.47E-04 \\
50  & 4.29E-04 & 4.50E-04 & 4.09E-04 \\
100 & 4.35E-04 & 4.48E-04 & 4.33E-04 \\
150 & 4.69E-04 & 4.36E-04 & 3.64E-04 \\
\bottomrule
\end{tabular}
\end{minipage}
\end{center}
\end{table}

Table \ref{table:table8} shows the $L_{2}$-error for different network architectures, while the number of initial points, boundary points, and collocation points is fixed to 100, 100, and 500, respectively, the $\lambda$ is fixed to 1 too. We observe that as the number of layers and neurons increases, the prediction becomes more accurate since the model becomes more powerful to approximate complex functions. While using too many layers and neurons can also cause problems such as overfitting, where the model is complex enough and the information contained in the training set is insufficient to train all the neurons, resulting in the degradation of performance.

\begin{table}[htbp]
\renewcommand{\arraystretch}{1.2}
\begin{center}
\begin{minipage}{250pt}
\caption{$L_{2}$-error under different choices of network architectures for advection equation}
\label{table:table8}
\begin{tabular}{cccc}
\toprule
\diagbox{Hidden layers}{Neurons}  & 20       & 30       & 40       \\
\midrule
2 & 4.55E-04 & 4.60E-04 & 4.36E-04 \\
3 & 6.83E-04 & 4.48E-04 & 6.70E-04 \\
4 & 5.43E-04 & 7.86E-04 & 7.31E-04 \\
\bottomrule
\end{tabular}
\end{minipage}
\end{center}
\end{table}

Table \ref{table:table9} represents the influence of the hyperparameter $\lambda$, while the number of initial points, boundary points, and collocation points is fixed to 100, 100, and 500, respectively, the architecture is 3 hidden layers with 30 neurons per layer. As expected, different $\lambda$ has a different influence on the accuracy of prediction.

\begin{table}[htbp]
\renewcommand{\arraystretch}{1.2}
\begin{center}
\begin{minipage}{400pt}
\caption{$L_{2}$-error under different choices of hyperparameter $\lambda$ for advection equation}
\label{table:table9}
\begin{tabular}{cccccccc}
\toprule
$\lambda$ & 0.01     & 0.05     & 0.1      & 0.5      & 1        &5        & 10       \\
\midrule
$L_{2}$   & 8.57E-04 & 7.74E-04 & 6.43E-04 & 4.78E-04 & 4.48E-04 &5.19E-04 & 5.49E-04 \\
\bottomrule
\end{tabular}
\end{minipage}
\end{center}
\end{table}

\section{Conclusion}
\label{sec:conclusion}
In this paper, we introduce Variance-Involved Physics-Informed Neural Networks, a novel framework considering the influence of variance. We use an explicit and efficient way to model uncertainty by viewing the observed value as a sample from a Gaussian distribution. And we introduce our idea as an auxiliary to assist vanilla PINNs to achieve a more accurate result at a faster speed of convergence. Five numerical examples are presented in order to investigate the performance of our proposed method. Our results show that VI-PINNs can improve the accuracy of PINNs and converge faster.

\section*{Declarations}
\textbf{Competing Interests} The authors have no competing interests to declare that are relevant to the content of this article.

\section*{Data availability}
The datasets generated during and/or analysed during the current study are available from the corresponding author on reasonable request.

\bibliography{refs}

\end{document}